# A Robust Probability-based Joint Registration Method of Multiple Point Clouds Considering Local Consistency

Ling-jie Su, Wei Xu*, *Member, IEEE*, Shu-yang Zhao, Yu-qi Cheng, and Wen-long Li, *Member, IEEE*

*Abstract*— In robotic inspection, joint registration of multiple point clouds is an essential technique for estimating the transformation relationships between measured parts, such as multiple blades in a propeller. However, the presence of noise and outliers in the data can significantly impair the registration performance by affecting the correctness of correspondences. To address this issue, we incorporate local consistency property into the probability-based joint registration method. Specifically, each measured point set is treated as a sample from an unknown Gaussian Mixture Model (GMM), and the registration problem is framed as estimating the probability model. By incorporating local consistency into the optimization process, we enhance the robustness and accuracy of the posterior distributions, which represent the one-to-all correspondences that directly determine the registration results. Effective closed-form solution for transformation and probability parameters are derived with Expectation-Maximization (EM) algorithm. Extensive experiments demonstrate that our method outperforms the existing methods, achieving high accuracy and robustness with the existence of noise and outliers. The code will be available at https://github.com/sulingjie/JPRLC_registration.

## I. INTRODUCTION

3D scanning technology has become essential due to high accuracy, non-contact measurement, and high efficiency [1]. In robotic inspection field, one critical task is to align the coordinates of the scanned point clouds for localization or quality assessment, known as the registration procedure. To estimate the multiple transformation relationships between the identical parts, such as the multiple blades in a propeller [2], joint registration is required to align point clouds simultaneously. Although several distance-based methods have been proposed to address joint registration task [3], [4], [5], their accuracy and robustness are unsatisfactory due to the sensitivity to noise and outliers.

Comparatively, probability-based registration methods are more applicable due to the stable one-to-all correspondence strategy. However, existing probability-based registration methods still tend to produce degenerate results when addressing defective data with noise and outliers [6]. The unsatisfactory results arise from the incorrect correspondences, a factor that directly determines the accuracy and robustness of the registration method. The unstable correspondences disturbed by noise and outliers directly reduce the registration performance. To overcome this limitation, we consider incorporating the local consistency property of probability model to generate more reliable correspondences, resulting in a robust and accurate registration method.

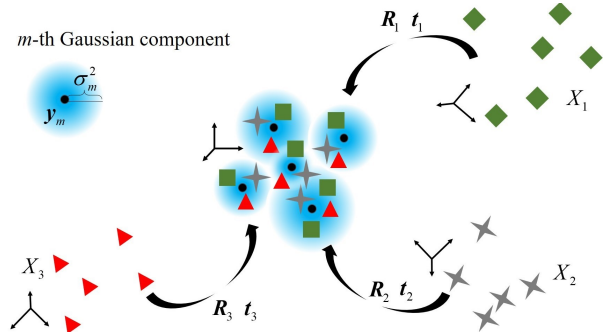

Fig. 1. Joint registration of multiple point clouds.

Specifically, we present a robust joint registration method within the probability-based framework. Each point set is treated as a sample from an unknown GMM, with both transformation and probability parameters optimized concurrently. To mitigate sensitivity to noise and outliers, we propose the local consistency constraint that enforces similarity in the posterior distributions among neighboring points, where the distributions represent the one-to-all correspondences that determine the registration result. By incorporating the constraint, the correspondences become more stable in the presence of noise and outliers, leading to more robust and accurate registration across point clouds. The proposed probability-based joint registration method is shown in Fig. 1. The contributions of our work are outlined below:

1) We propose local consistency to solve the key problem in joint registration, the mismatching of the corresponding points, thereby improving the robustness and accuracy of joint registration method, significantly enhancing the ability to handle noise and outliers.

2) The effective closed-form solutions considering local consistency are derived under EM framework to estimate the optimal transformation and probability parameters, achieving the probability model fitting task.

3) Extensive experiments have been conducted to verify the performance of the proposed method. The experiment results demonstrate effectiveness of local consistency and our method's superiority over the existing methods.

### A. Related Work

In this section, we review the existing methods closely related to our work, including pairwise registration and joint registration

*1) Pairwise Registration*

The pairwise registration methods are divided into three categories, including distance-based, feature-based, and probability-based methods. Iterative Closest Point (ICP) [7], the first distance-based registration method, takes the closest points as correspondences and minimizes the distances

*Research supported by the National Natural Science Foundation of China under Grant 52205524, Grant 52188102, and Grant 52075203. (Corresponding author: Wei Xu.)*

Ling-jie Su, Wei Xu, Shu-yang Zhao, Yu-qi Cheng, and Wen-long Li are with the State Key Laboratory of Intelligent Manufacturing Equipment and Technology, Huazhong University of Science and Technology, Wuhan 430074, China (e-mail: ljsu@hust.edu.cn, weixu.chn@gmail.com syzhao@hust.edu.cn, yuqicheng@hust.edu.cn, wlli@mail.hust.edu.cn).

between correspondences. However, the one-to-one correspondence strategy is sensitive to noise and outliers, reducing the registration performance. Following ICP, several works designed constraints to prohibit pose updating along degenerate directions with defective data [8], [9], [10]. Feature-based methods match the key points extracted from point clouds as correspondences to estimate the optimal transformation between point clouds [11], [12], [13]. To obtain general features, Some researchers utilize neural networks to extract point cloud features [14], [15], [16]. Different from the above methods, the probability-based methods employ the more stable one-to-all correspondence strategy, casting the registration problem into a probability model fitting task. Specifically, points in one set are treated as centroids of GMM components, while points in the other set are considered as samples. The transformation and probability parameters are estimated simultaneously within EM framework [17], [18].

However, when facing the joint registration problem of multiple point clouds, the pairwise registration methods cannot estimate the poses of all point clouds simultaneously due to cumulative errors and bias.

*2) Joint Registration*

The joint registration methods can generally be classified into two categories: distance-based and probability-based approaches.

Distance-based methods focus on refining point cloud coordinates by optimizing a predefined distance metric, using the one-to-one correspondence strategy. Bergevin et al. [3] first extended the ICP method to multiple registration situation, considering all views as a whole and minimizing the closest distances of all correspondences simultaneously. Different local features and optimization strategies were considered to improve the registration performance [4], [19], [20], [21]. Additionally, Cheng et al. [5] considered the distances and variances between corresponding points as the objective function to be optimized. The added variance constraint improves the registration performance regarding data with uneven density.

Compared with the sensitivity to noise and outliers in one-to-one correspondence strategy, the one-to-all strategy in probability-based methods is more stable in real applications. Evangelidis et al. [22] proposed JRMPC method that takes each point cloud as a sample from an unknown GMM. The transformation and probability parameters are optimized with EM algorithm simultaneously. In this way, there is no bias toward certain point set. Zhu et al. [23] cast the joint registration problem into a clustering task and solved the problem based on K-means. Min et al. [24] incorporated the orientational information into the registration procedure, representing the normal vectors with von Mises-Fisher distribution. Ma et al. [25] replaced GMM with Student's-*t* mixture model (SMM), which renders the method inherently robust to outliers and heavy-tail noise. Despite the high performance, the one-to-all correspondences can still be affected by noise and outliers, undermining the potential of the probability-based methods. To address the problem, we impose local consistency constraint on the correspondences, ensuring that neighboring points exhibit similar posterior distributions. The added constraint enhances both the robustness and accuracy of the joint registration method, particularly in the presence of noise and outliers.

## II. METHOD

In this section, we first review the joint registration method with native GMM. Then, the detailed form of local consistency property is proposed and incorporated into the optimization procedure. Finally, the closed-form solutions are derived under EM framework.

### A. Joint registration with native GMM

Given a point cloud set $X = [X_1, X_2, ..., X_N]$, where $N$ is the number of the set, the joint registration problem is to estimate rotation matrix $\boldsymbol{R}_j$ and translation vector $\boldsymbol{t}_j$ for each point cloud to align the coordinates together. Under the probability framework, the transformed point cloud can be regarded as samples from the Gaussian Mixture Model (GMM). The centroids of GMM components are $[y_1, y_2, ..., y_M]$, the variances are $[\sigma_1^2, \sigma_2^2, ..., \sigma_M^2]$, and the weights are $[\pi_1, \pi_2, ..., \pi_M]$, where $M$ is the number of the components. The $i$-th point in the $j$-th point cloud is denoted as $\boldsymbol{x}_{ji}$ and the probability of the transformed point $\boldsymbol{x}_{ji}$ being sampled is

$$p(\boldsymbol{x}_{ji}) = \sum_m \pi_m p(\boldsymbol{x}_{ji} | z_{ji} = m; \Theta) + \pi_{M+1} \mathcal{U}(\mathrm{V}) \quad (1)$$

where $p(\boldsymbol{x}_{ji} | z_{ji} = m; \Theta)$ is the probability of the transformed $\boldsymbol{x}_{ji}$ being sampled from the $m$-th Gaussian distribution and the specific form is as follows:

$$p(\boldsymbol{x}_{ji} | z_{ji} = m; \Theta) = (2\pi\sigma_m^2)^{-\frac{3}{2}} e^{-\frac{1}{2\sigma_m^2} \|\phi_j(\boldsymbol{x}_{ji}) - \boldsymbol{y}_m\|^2} \quad (2)$$

where $\Theta = \left\{ (\boldsymbol{R}_j, \boldsymbol{t}_j)_{j=1}^N, (\boldsymbol{y}_m, \sigma_m^2, \pi_m)_{m=1}^M \right\}$ is the parameter set. $\mathcal{U}(\mathrm{V}) = 1/\mathrm{V}$ is a uniform distribution, representing the outliers, and $\pi_{M+1}$ is the corresponding weight given in advance. $\phi_j(\boldsymbol{x}_{ji}) = \boldsymbol{R}_j \boldsymbol{x}_{ji} + \boldsymbol{t}_j$ represents the transformation of the $j$-th point cloud. $z_{ji}$ is the hidden variable corresponding with $\boldsymbol{x}_{ji}$ and $z_{ji} = m$ means that $\boldsymbol{x}_{ji}$ is sampled from the $m$-th Gaussian component. The hidden variable set is denoted as $\mathcal{Z}$.

The parameters are estimated under EM framework, where the expectation of the negative complete-data log-likelihood is minimized as follows:

$$\begin{aligned} Q_{\mathrm{GMM}}(\Theta | X, \mathcal{Z}) &= -E_{\mathcal{Z}} \left[ \log P(X, \mathcal{Z}; \Theta | X) \right] \\ &= -\sum_{\mathcal{Z}} P(\mathcal{Z} | X; \Theta) \log P(X, \mathcal{Z}; \Theta). \end{aligned} \quad (3)$$

Ignoring constants, Eq (3) can be expanded as follows:

$$Q_{\mathrm{GMM}}(\Theta) = \sum_{j,i,m} p_{jim} \frac{\|\phi_j(\boldsymbol{x}_{ji}) - \boldsymbol{y}_m\|^2}{2\sigma_m^2} + \frac{3}{2} \sum_{j,i,m} p_{jim} \log \sigma_m^2 - \sum_{j,i,m} p_{jim} \log \pi_m \quad (4)$$

where $p_{jim}$ represents the posterior $p(z_{ji} = m | \boldsymbol{x}_{ji}; \Theta^{\mathrm{old}})$. In the optimization process under EM framework, the posterior distributions are estimated in the E-step and are constant in the M-step. The detailed form of $p_{jim}$ is obtained with Bayesian

**Algorithm 1:** Joint Probability-based Registration with Local Consistency (JPRLC)
---
1: **Initialization:** Number of GMM components $M$, initialized parameter set $\Theta^0$, uniform distribution weight $\pi_{M+1}$, iteration number $D$.
2: Current iteration count $q = 1$.
3: **if** $q < D$
4:  E-step: Estimate posterior distributions $p_{jim}^q$ in (5) with $\Theta^{q-1}$.
   M-step:
5:  a). Estimate translation vector $t_j^q$ in (12) with $p_{jim}^q$, $R_j^{q-1}$, $y_m^{q-1}$, $(\sigma_m^2)^{q-1}$, and $\pi_m^{q-1}$.
6:  b). Estimate rotation matrix $R_j^q$ in (17) with $p_{jim}^q$, $t_j^q$, $y_m^{q-1}$, $(\sigma_m^2)^{q-1}$, and $\pi_m^{q-1}$.
7:  c). Estimate centroids of Gaussian components $y_m^q$ in (19) with $p_{jim}^q$, $t_j^q$, $R_j^q$, $(\sigma_m^2)^{q-1}$, and $\pi_m^{q-1}$.
8:  d). Estimate variances of Gaussian components $(\sigma_m^2)^q$ in (20) with $p_{jim}^q$, $t_j^q$, $R_j^q$, $y_m^q$, and $\pi_m^{q-1}$.
8:  e). Estimate weights of Gaussian components $\pi_m^q$ with $p_{jim}^q$, $t_j^q$, $R_j^q$, $y_m^q$, and $(\sigma_m^2)^q$.
8: $q = q + 1$.
9: **end**
10: **return:** $\{R_j^*, t_j^*\}_{j=1}^N$.

formula as follows:
$$\begin{aligned}p_{jim} &= p(z_{ji} = m \mid x_{ji}; \Theta^{old}) \\ &= \frac{p(z_{ji} = m) p(x_{ji} \mid z_{ji} = m; \Theta^{old})}{p(x_{ji})} \\ &= \frac{\pi_m (2\pi\sigma_m^2)^{-\frac{3}{2}} e^{-\frac{1}{2\sigma_m^2}\|\phi_j(x_{ji}) - y_m\|^2}}{\sum_m \pi_m (2\pi\sigma_m^2)^{-\frac{3}{2}} e^{-\frac{1}{2\sigma_m^2}\|\phi_j(x_{ji}) - y_m\|^2} + \pi_{M+1}/V}.\end{aligned} \quad (5)$$

By casting the joint registration problem into a GMM fitting task, the coordinates of all point clouds can be aligned simultaneously with no bias.

*B. Joint registration with local consistency constraint*

In the native GMM, the one-to-all correspondences can be disturbed by noise and outliers. The defective data is common in complex industrial measurement scenarios, thereby decreasing the robustness and accuracy of the joint registration method. To overcome this problem, additional constraints to correspondences are indispensable to improve the robustness against defective data, including noise and outliers. In this paper, we introduce local consistency property of probability model to constrain the posterior distributions, improving the stability of correspondences, thereby enhancing the registration performance. As described in [26], the posterior distributions within neighboring points are similar. It is intuitive that when two points are close, it is likely that they are sampled from the same Gaussian component. Fig. 2 illustrates the posterior distributions of neighboring points.

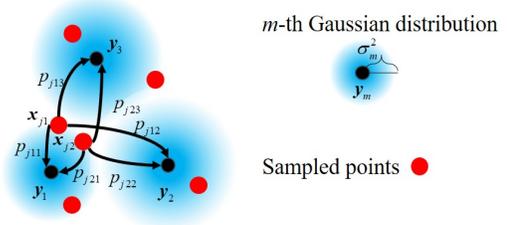

Fig 2. Posterior distributions of two neighboring points.

On one side, the posteriors of inlier points are given higher weights in the optimization process with the added constraint. In contrast, invalid correspondences with outliers are weighted less because outliers are sparser and less likely to have neighbors, thereby mitigating the negative influence of outliers. On the other side, the posteriors of local points are considered collectively, enhancing the method's robustness against disturbances caused by noise. KL-Divergence is used to measure the similarity between distributions. Given two posterior distributions $p(z_{ji} \mid x_{ji})$ and $p(z_{jb} \mid x_{jb})$, the similarity between them is defined as follows:

$$D_{jib} = \frac{1}{2}\{D[p(z_{ji} \mid x_{ji}) \| p(z_{jb} \mid x_{jb})] + D[p(z_{jb} \mid x_{jb}) \| p(z_{ji} \mid x_{ji})]\} \quad (6)$$

where
$$\begin{aligned}&D[p(z_{ji} \mid x_{ji}) \| p(z_{jb} \mid x_{jb})] \\ &= \sum_m p(z_{ji} = m \mid x_{ji}) \log \frac{p(z_{ji} = m \mid x_{ji})}{p(z_{jb} = m \mid x_{jb})}.\end{aligned} \quad (7)$$

Thus, the local consistency constraint is defined as follows:
$$Q_{LC}(\Theta) = \sum_{j,i,b} w_{jib} D_{jib} \quad (8)$$

where $w_{jib}$ represents whether $x_{ji}$ and $x_{jb}$ are neighboring points. When $x_{ji}$ and $x_{jb}$ are neighbors, $w_{jib} = 1$, else $w_{jib} = 0$. Combining (5) and (7), we can get the specific form of the similarity of posterior distributions as follows:

$$D_{jib} = \sum_m \frac{p_{jim} - p_{jbm}}{4\sigma_m^2} \left( \|\phi_j(x_{jb}) - y_m\|^2 - \|\phi_j(x_{ji}) - y_m\|^2 \right). \quad (9)$$

The objective function considering local consistency for minimization is denoted as follows:
$$Q(\Theta) = Q_{GMM}(\Theta) + \lambda Q_{LC}(\Theta) \quad (10)$$

where $\lambda$ is used to control the weight of local consistency term and is given in advance.

*C. Closed-form solution*

To estimate effective values of $\Theta$ minimizing (10), the closed-form solutions considering local consistency term are derived in M-step under EM framework. Throughout the M-step, the posterior distributions remain constant and the values are estimated in the E-step in advance.

Ignoring the constants relating to $t_j$, the objective function is as follows:

$$Q(t_j) = \sum_{i,m} \frac{p_{jim}}{2\sigma_m^2}\left(2x_{ji}^T R_j^T t_j - 2y_m^T t_j + t_j^T t_j\right)$$
$$+ \frac{\lambda}{2}\sum_{i,b,m} w_{jib}\frac{p_{jim}-p_{jbm}}{\sigma_m^2}\left(x_{jb}-x_{ji}\right)^T R_j^T t_j. \tag{11}$$

The optimal translation vector $t_j^*$ is obtained by solving the function $\partial Q(t_j)/\partial t_j = 0$. The result is as follows:

$$t_j^* = \frac{\mu_{yj}}{N_{pj}} - R_j \frac{\mu_{xj}}{N_{pj}}$$
$$\mu_{xj} = \sum_{i,m}\frac{p_{jim}}{\sigma_m^2}x_{ji} + \frac{\lambda}{2}\sum_{i,b,m}w_{jib}\frac{p_{jim}-p_{jbm}}{\sigma_m^2}\left(x_{jb}-x_{ji}\right) \tag{12}$$
$$\mu_{yj} = \sum_{i,m}\frac{p_{jim}}{\sigma_m^2}y_m, \quad N_{pj} = \sum_{i,m}\frac{p_{jim}}{\sigma_m^2}.$$

Similarly, ignoring the constants relating to $R_j$ and supplanting $t_j$ with $t_j^*$, the objective function is as follows:

$$Q(R_j) = -\sum_{i,m}\frac{p_{jim}}{\sigma_m^2}y_{mj}'^T R_j x_{ji}'$$
$$+ \frac{\lambda}{2}\sum_{i,b,m}w_{jib}\frac{p_{jim}-p_{jbm}}{\sigma_m^2}y_{mj}'^T R_j\left(x_{ji}'-x_{jb}'\right) \tag{13}$$
$$x_{ji}' = x_{ji} - \frac{\mu_{xj}}{N_{pj}}$$
$$y_{mj}' = y_m - \frac{\mu_{yj}}{N_{pj}}.$$

The optimal rotation matrix $R_j^*$ is estimated by minimizing $Q(R_j)$ as follows:

$$R_j^* = \arg\min Q(R_j)$$
$$= \arg\max \underbrace{\sum_{i,m}\frac{p_{jim}}{\sigma_m^2}y_{mj}'^T R_j x_{ji}'}_{Q_1}$$
$$+ \underbrace{\frac{\lambda}{2}\sum_{i,b,m}w_{jib}\frac{p_{jbm}-p_{jim}}{\sigma_m^2}y_{mj}'^T R_j\left(x_{ji}'-x_{jb}'\right)}_{Q_2} \tag{14}$$
$$= \arg\max(Q_1 + Q_2).$$

Utilizing the properties $\text{Tr}(\mathbf{ABC}) = \text{Tr}(\mathbf{CAB})$, $Q_1$ and $Q_2$ can be represented as follows:

$$Q_1 = \text{Tr}\left(R_j \underbrace{\sum_{i,m}\frac{p_{jim}}{\sigma_m^2}x_{ji}'y_{mj}'^T}_{\mathbf{H}_1}\right)$$
$$Q_2 = \text{Tr}\left(R_j \underbrace{\frac{\lambda}{2}\sum_{i,b,m}w_{jib}\frac{p_{jbm}-p_{jim}}{\sigma_m^2}\left(x_{ji}'-x_{jb}'\right)y_{mj}'^T}_{\mathbf{H}_2}\right). \tag{15}$$

Thus, the optimal rotation matrix can be obtained as follows:

$$R_j^* = \arg\max\left(\text{Tr}(R_j\mathbf{H}_1) + \text{Tr}(R_j\mathbf{H}_2)\right)$$
$$= \arg\max \text{Tr}(R_j\mathbf{H}) \tag{16}$$

where $\mathbf{H} = \mathbf{H}_1 + \mathbf{H}_2$. The maximized orthonormal matrix $\mathbf{A}$ for $\text{Tr}(\mathbf{AH})$ is $\mathbf{A} = \mathbf{VU}^T$, where $\mathbf{V}$ and $\mathbf{U}$ are obtained by singular value decomposition of $\mathbf{H}$, $\mathbf{H} = \mathbf{USV}^T$. Thus, the optimal rotation matrix is as follows:

$$R_j^* = \mathbf{V}\text{diag}\left(\begin{bmatrix}1 & 1 & \det(\mathbf{VU}^T)\end{bmatrix}\right)\mathbf{U}^T. \tag{17}$$

Substituting the optimal translation vector and rotation matrix into the objective function and ignoring the constants, the objective function relating to $y_m$ is as follows:

$$Q(y_m) = \sum_{j,i}\frac{p_{jim}}{2\sigma_m^2}\left(-2x_{ji}^T R_j^T y_m - 2t_j^T y_m + y_m^T y_m\right)$$
$$+ \frac{\lambda}{2}\sum_{j,i,b}w_{jib}\frac{p_{jim}-p_{jbm}}{\sigma_m^2}\left(x_{ji}^T - x_{jb}^T\right)R_j^T y_m. \tag{18}$$

By solving the equation $\partial Q(y_m)/\partial y_m = 0$, the optimal Gaussian centroid is obtained as follows:

$$y_m^* = \frac{\sum_{j,i}\frac{p_{jim}}{\sigma_m^2}\left(R_j x_{ji} + t_j\right)}{\sum_{j,i}\frac{p_{jim}}{\sigma_m^2}}$$
$$- \frac{\frac{\lambda}{2}\sum_{j,i,b}w_{jib}\frac{p_{jim}-p_{jbm}}{\sigma_m^2}R_j\left(x_{ji}-x_{jb}\right)}{\sum_{j,i}\frac{p_{jim}}{\sigma_m^2}}. \tag{19}$$

Similarly, the optimal variances are obtained by solving the equation $\partial Q(\Theta)/\partial \sigma_m^2 = 0$. The result is as follows:

$$\sigma_m^{2*} = \frac{1}{3\sum_{j,i}p_{jim}}\sum_{j,i}p_{jim}\left\|\phi_j(x_{ji}) - y_m\right\|^2$$
$$+ \frac{\lambda}{6\sum_{j,i}p_{jim}}\sum_{j,i,b}w_{jib}\left(p_{jim}-p_{jbm}\right)\times \tag{20}$$
$$\left(\left\|\phi_j(x_{jb})-y_m\right\|^2 - \left\|\phi_j(x_{ji})-y_m\right\|^2\right)$$

As for the weight of Gaussian component, the objective function relating to $\pi_m$ is as follows:

$$Q(\pi_1,\pi_2,\ldots,\pi_m) = -\sum_{j,i,m}p_{jim}\log\pi_m$$
$$\text{s.t.} \sum_m \pi_m = 1 - \pi_{M+1}. \tag{21}$$

To solve the problem, the Lagrange multiplier method is utilized and the optimal weight $\pi_m^*$ is as follows:

$$\pi_m^* = (1-\pi_{M+1})\frac{\sum_{j,i}p_{jim}}{\sum_{j,i,m}p_{jim}} \tag{22}$$

The workflow of the proposed method is presented in Algorithm 1.

## III. EXPERIMENTS

In this section, we evaluate the performance of the proposed method, referred to as *joint probability-based registration with local consistency* (JPRLC), against widely-used registration methods on measured blade point cloud data. First, to assess the pairwise registration performance, we compare JPRLC with ICP [7], ECMPR [18], and JRMPC [22] on pairwise registration task. Next, to demonstrate the effectiveness of the local consistency constraint, we adjust the weight $\lambda$ and repeat the experiments on a set of point clouds. Finally, to evaluate the robustness and accuracy on defective data with noise and outliers, we conduct experiments on data with varying levels of noise and outliers, comparing JPRLC with state-of-the-art methods, including MICP [3], JRMPC [22], KmeansReg [23], EMPTM [25], and MVGR [5]. To verify the registration performance quantitively, we fix the first point cloud and calculate the error of the rest with Rooted Mean Squared Error (RMSE), which is defined as follows:

$$\text{RMSE} = \sqrt{\frac{1}{\sum_{j=2}^{N} N_j} \sum_{j=2}^{N} \sum_{i=1}^{N_j} \left\| \phi_{\text{cal}}^j(\boldsymbol{x}_{ji}) - \phi_{\text{gt}}^j(\boldsymbol{x}_{ji}) \right\|^2} \quad (23)$$

where $\phi_{\text{cal}}^j(\boldsymbol{x}_{ji})$ and $\phi_{\text{gt}}^j(\boldsymbol{x}_{ji})$ are calculated and ground-truth transformation of point $\boldsymbol{x}_{ji}$ under the coordinate of the first point cloud.

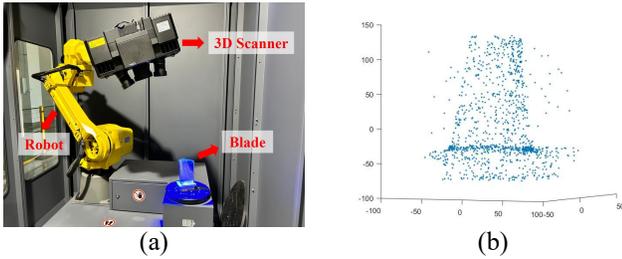

| (a) | (b) |

Fig. 3. (a) Robotic measurement scene; (b) Blade point cloud with added noise and outliers.

### A. Initialization and Data Preparation

The initialized parameters play an important role in optimization procedure, the parameters in JPRLC needing initialization include $\boldsymbol{t}_j$, $\boldsymbol{R}_j$, $\boldsymbol{y}_m$, $\sigma_m^2$, and $\pi_m$. For transformation parameters, the translation vector is initialized by the center differences between the point cloud and GMM, expressed $\boldsymbol{t}_j^0 = \bar{\boldsymbol{\mu}}_Y - \bar{\boldsymbol{\mu}}_{X_j}$, where $\bar{\boldsymbol{\mu}}_Y$ represents the center of GMM and $\bar{\boldsymbol{\mu}}_{X_j}$ represents the center of point cloud $X_j$. Rotation matrix is initialized by an identical matrix, expressed as $\boldsymbol{R}_j^0 = \boldsymbol{I}_3$. For GMM parameters, the way to initialize $\boldsymbol{y}_m$ is to distribute it on the surface of a sphere, whose radius is half of the convex sphere of all point clouds and the center $\bar{\boldsymbol{\mu}}_Y$ is fixed at the centroid of all point clouds. The weight of uniform distribution $\pi_{M+1}$ is set to be 0.1. The number of Gaussian components $M$ is set to be 1,000. The variance $\sigma_m^2$ in GMM is empirically initialized to be 1,000 and the weight of each component in GMM is initialized to be $(1-\pi_{M+1})/M$. For each method, the iteration number is fixed at 100.

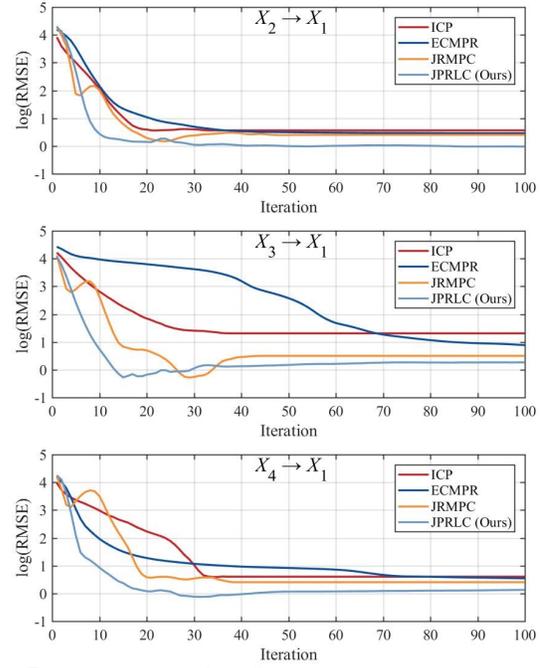

Fig. 4. Registration results of pairwise registration.

The blade data in the experiments are measured by ZEISS GOM ATOS SCANBOX 5,108, which is known for high accuracy. The measurement scene is shown in Fig 3 (a). Point cloud $X_1$ is down-sampled from the original data with 1,000 points. Point cloud $X_2$, $X_3$, and $X_4$ are sampled randomly from $X_1$, and the sizes are 700, 500, and 300 separately. To verify the robustness and accuracy of JPRLC, Gaussian noise with different standard deviations and outliers with different ratios are added in each trial. The rotation angle along each axis is generated within $[-60°, 60°]$ and the translation distance along each axis is within $[-40, 40]$. One case of the generated data is shown in Fig. 3 (b).

### B. Pairwise Registration Experiment

To verify the performance on pairwise registration, a special case of joint registration, we conduct the pairwise registration experiment on defective data. Gaussian noise with a standard deviation of 4.0 and outliers at a ratio of 10% are added to $X_2$, $X_3$, and $X_4$. Then, the registrations with $X_1$ are conducted separately. $\lambda$ is fixed at 0.1. Fig. 4 shows the log(RMSE) over 100 iterations of each method.

From the results in Fig. 4, JPRLC consistently achieves the lowest registration error across all trials. In contrast, ICP relies on one-to-one correspondence approach, making it highly susceptible to noise and outliers. Incorrect correspondences in ICP can significantly reduce registration accuracy. ECMPR and JRMPC are probability-based methods that adopt the one-to-all correspondence strategy. However, noise and outliers can degrade the correctness of the correspondences and decrease the registration accuracy. JPRLC addresses this limitation by incorporating the local consistency property of GMM, constraining the posterior distributions, thereby mitigating the negative impact of noise and outliers. The experiment results validate the effectiveness of JPRLC in improving pairwise registration accuracy.

## C. Local Consistency Effectiveness Experiment

In this experiment, to verify the effectiveness of the proposed local consistency, the weight $\lambda$ is adjusted and the experiments are repeated to estimate the transformation relationships among $X_1$, $X_2$, $X_3$, and $X_4$ simultaneously. Gaussian noise with a 3.0 standard deviation and 10% outliers are added to each set. For each parameter, 5 trials are repeated. The registration results are shown in Fig. 5.

The experiment results show that the registration accuracy improves when the local consistency term is introduced, demonstrating the effectiveness of the added constraint. The local consistency constrains the posterior distributions among neighboring points, enhancing the correctness of correspondences in the presence of noise and outliers. When $\lambda$ is 0, the method deteriorates into native GMM-based approach. When $\lambda$ is around 0.1, JPRLC achieves optimal registration performance for this set of data. When $\lambda$ is over 0.1, the accuracy decreases because the excessive weight of local consistency impairs the maximization of the log-likelihood expectation.

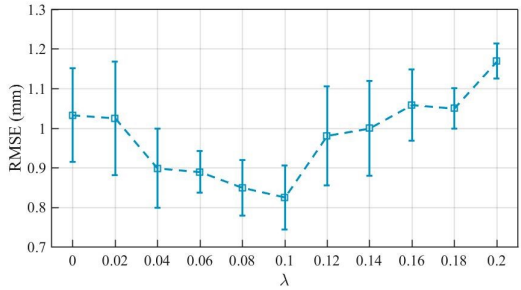

Fig. 5. Registration results of local consistency effectiveness experiment.

## D. Noise and Outlier Robustness Experiment

In this experiment, we evaluate the robustness and accuracy of JPRLC with the existence of different levels of noise and outliers. First, we test the data added Gaussian noise with different standard deviations, ranging from 1.0 to 5.0 with an interval of 0.5, while keeping the outlier ratio fixed at 10% and $\lambda$ is fixed at 0.1. Then, we test the data added outliers with different ratios, ranging from 5% to 50% with an interval of 5%. The standard deviation of Gaussian noise is fixed at 3.0 and $\lambda$ is fixed at 0.1. Each experiment is repeated for 12 times, with the noise and outliers regenerated for each trial and the ground-truth transformation randomly selected. To avoid misalignments in the registration results, a trial is considered successful if the final RMSE is less than 10mm. The success rates of each method under different noise and outlier levels are presented in Fig. 6. The mean values and standard deviation ranges of registration errors for successful trials are shown in Fig. 7 and Fig. 8. The absence of EMPTR when the outlier ratio is 50% indicates no successful trail and the missing error bars indicate only one successful trial.

For distance-based methods, MICP [3] and MVGR [5], one-to-one correspondences are highly susceptible to noise and outliers, often leading to registration failure and reduced accuracy. In the KmeansReg method [23], noise-induced disturbance on points can misclassify points into incorrect clusters, while outliers tend to be mixed with inliers in clusters, further contributing to registration failure and diminishing accuracy. For JRMPC [22] and EMPTM [25], the lack of constrained posteriors increases the instability of corresponding points, decreasing accuracy with the existence of noise and outliers. In contrast, JPRLC leverages the local consistency term to constrain the posterior distributions, representing the one-to-all correspondences. The correctness of correspondences directly determines the registration accuracy, allowing JPRLC to maintain stable performance with the existence of noise and outliers. The proposed JPRLC method outperforms other methods in terms of both success rate and registration accuracy.

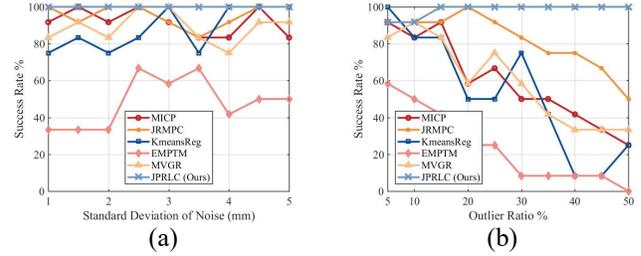

(a)          (b)

Fig. 6. Success rate of robustness experiment. (a) Results with different levels of noise; (b) Results with different levels of outliers.

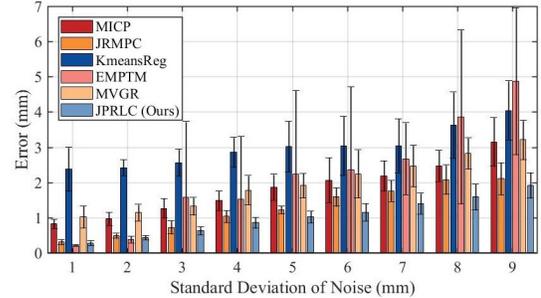

Fig 7. Mean RMSE with different levels of noise.

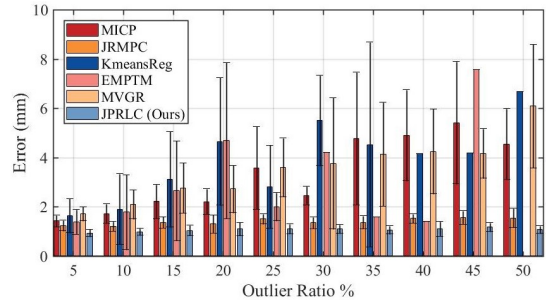

Fig 8. Mean RMSE with different levels of outliers.

## IV. CONCLUSION

To address the challenge of joint registration with defective data, including noise and outliers, we incorporate local consistency property into probability-based registration method, constraining the posteriors among neighboring points, where posteriors represent the one-to-all correspondences that determine the registration results. Effective closed-form solutions of transformation and probability parameters are provided under EM framework. Extensive experiments are conducted to evaluate the robustness and accuracy of the proposed method, comparing its performance with state-of-the-art registration techniques. The results demonstrate the effectiveness of local consistency and show that JPRLC outperforms the existing methods in the presence of noise and outliers.